\documentclass{llncs}
\pdfoutput=1
\usepackage{booktabs} 
\usepackage{subdepth}
\usepackage{bm}
\usepackage{dsfont}
\usepackage{xspace}
\usepackage{algorithm}
\usepackage{amsmath}
\usepackage{amsfonts}
\usepackage{graphicx}
\usepackage{cleveref}
\usepackage{pdfpages}
\usepackage[noend]{algpseudocode}
\usepackage{todonotes}
\usepackage{paralist}
\usepackage{pgf}

\newcommand{\HD}{\ensuremath{{\text{HD}}}\xspace}

\newcommand{\ary}{\ensuremath{{\text{arity}}}\xspace}

\newcommand{\cor}{\ensuremath{{\text{cor}}}\xspace}

\newcommand{\fkern}{\ensuremath{{\text{k}}}\xspace}
\newcommand{\fdistPhD}{\ensuremath{{\text{d}_\text{PhD}}}\xspace} 
\newcommand{\fdistTED}{\ensuremath{{\text{d}_\text{TED}}}\xspace} 
\newcommand{\fdistSHDa}{\ensuremath{{\text{d}_\text{SHD1}}}\xspace} 
\newcommand{\fdistSHDb}{\ensuremath{{\text{d}_\text{SHD2}}}\xspace} 

\DeclareMathOperator*{\argmin}{arg\,min}                   

\usepackage{bbding} 
\begin{document}
\title{Linear Combination of Distance Measures for Surrogate Models in Genetic Programming\thanks{The final authenticated version of this publication will appear in the proceedings of the 15th International Conference on
Parallel Problem Solving from Nature 2018 (PPSN XV), published in the LNCS by Springer}}

\author{Martin Zaefferer \Envelope \inst{1} \and J\"org Stork\inst{1} \and Oliver Flasch\inst{2}  \and Thomas Bartz-Beielstein\inst{1}}

\institute{
TH K\"oln, Institute of Data Science, Engineering, and Analytics, \\
Steinm\"ullerallee 6, 51643~Gummersbach, Germany, \email{martin.zaefferer@th-koeln.de},
\email{joerg.stork@th-koeln.de}, \email{thomas.bartz-beielstein@th-koeln.de}
\and
sourcewerk GmbH, 
Roseggerstra{\ss}e 59,
44137 Dortmund, Germany\\
\email{oliver.flasch@sourcewerk.de}
}

\maketitle

\begin{abstract}
Surrogate models are a well established approach to
reduce the number of expensive function evaluations in continuous optimization.
In the context of genetic programming, surrogate modeling
still poses a challenge, due to the complex genotype-phenotype
relationships. 
We investigate how different genotypic and phenotypic distance measures can be used to learn Kriging models as surrogates. 
We compare the measures and suggest to use their linear combination in a kernel. 

We test the resulting model in an optimization framework, using symbolic regression problem instances as a benchmark.
Our experiments show that the model provides valuable information. 
Firstly, the model enables an improved optimization performance compared
to a model-free algorithm. 
Furthermore, the model provides
information on the contribution of different distance measures.
The data indicates that a phenotypic distance measure is important during the early stages of an optimization run when less data is available. 
In contrast, genotypic measures, such as the tree edit distance, contribute more during the later stages.
\end{abstract}
\section{Introduction}\label{sec:intro}
Genetic programming (GP) automatically evolves computer programs that aim to solve a task.
This idea goes back to fundamental work by John Koza~\cite{koza1994genetic} and follows the principles of evolutionary computation. 
The computer programs are individuals subject to an evolutionary process, which improves them based on their fitness, i.e., their ability to solve a problem. 
Examples for GP tasks are symbolic regression (SR), classification, and production scheduling~\cite{Flasch2015b,Nguyen2017}. 

Expensive fitness functions pose a challenge to evolutionary algorithms, including GP.
This occurs, e.g., when the fitness function requires laboratory experiments or extensive simulations.
Frequently, Surrogate Model-Based Optimization 
(SMBO) is used to deal with expensive evaluations~\cite{Bart16n}.
Most SMBO research focuses on problems with continuous variables, where many competitive regression models are available. 
In the context of GP, the use of surrogates is not well researched. 
This might seem surprising, as the computational bottleneck of most GP applications is the evaluation of fitness cases. 
Unfortunately, surrogate modeling of GP tasks, such as SR, is difficult, because it subsumes modeling of a complex genotype-phenotype-fitness mapping. 
Recent work in deep learning suggests that this mapping can be approximated, at least in certain domains of program synthesis~\cite{Parisotto2016}. 

In the last years, combinatorial search spaces were treated successfully with SMBO, by using distance-based models~\cite{Moraglio2011,Zaefferer2014b}.
However, there is no generic way for choosing an adequate distance measure. 
For complex tree shaped structures, which occur in GP, it is challenging to select a suitable distance measure and find a feasible modeling approach.
For that reason, we will focus on the following research questions regarding SMBO for GP and tree-shaped structures:
\begin{compactenum}
\item How do different distance measures compare to each other?
\item What impact do these distances have on the model?
\item How does SMBO based on a linear combination of these distances compare to a model-free Evolutionary Algorithm (EA) and random search?
\end{compactenum}
To answer these questions, we will utilize bi-level optimization problems based on different SR tasks as test functions. 
While these test functions are not that expensive to evaluate (and hence are not a natural use-case for surrogate models), they present a challenging benchmark for the proposed models. They allow us to gain insights into the topics summarized by our research questions. We expect that our result can be transferred to other problems with tree shaped structures, such as program synthesis for general purpose or domain-specific languages. 

%
%
%

\section{Related Work}
In the following, we will differentiate between two approaches, which will be further referred to as a) SMBO and b) SAEA.
\begin{compactenum}[a)]
\item Sequential SMBO
generates new candidate solutions by performing a search procedure on
the surrogate model, e.g., as described for the Efficient Global Optimization (EGO) algorithm by Jones et al.~\cite{Jones1998}. 
\item Approaches that utilize surrogates to assist an EA (SAEA), e.g., as described by Jin~\cite{Jin2011}. For example, the surrogate is utilized to support the selection process of an EA by predicting the fitness of proposed offspring. 
\end{compactenum}


Most studies on GP and surrogate modeling focus on SAEA.
Kattan and Ong~\cite{Kattan2014} describe an SAEA approach with two distinct Radial Basis Function Network (RBFN) models (semantic and fitness). The conjunction of both models is used to evolve a subset of the population. 
They report superiority of their approach over standard GP for three different tasks, including SR.

Hildebrandt and Branke~\cite{Hildebrandt2014} present a phenotypic distance. They optimize job dispatching rules with an SAEA approach. Their surrogate model is a nearest neighbor regression model based on the phenotypic distance. 
They demonstrate that their model allows for a faster evolution of good solutions. 
This approach is also discussed and extended by Nguyen et al.~\cite{Nguyen2014,Nguyen2016}. 


To the best of our knowledge, only Moraglio and Kattan~\cite{Moraglio2011b} describe an SMBO approach to GP where a very limited number of function evaluations is allowed.
They use an RBFN with appropriate distance measures. 
Their results did not indicate a significant improvement over the use of a model-free optimization approach.

In contrast to these works, we aim to learn Kriging models (following the idea of EGO~\cite{Jones1998}) and employ them in an SMBO framework with a severely limited number of 100 fitness function evaluations. Our models are based on a linear combination of three diverse distances. 
Like several of the above described studies, we use SR as a test case. 
We want to show that
the relation between complex structures and their associated fitness can be learned and exploited for optimization purposes.
Although SR is not particularly expensive, we argue that it presents a difficult and challenging test case to investigate whether our proposed models are able to learn such a complex search landscape.

\section{A Test Case for SMBO-GP: Bi-level Symbolic Regression}
In SR, a regression task is solved by evolving symbolic expressions. In essence, SR searches for a formula that best represents a given data set.
The formulas can be represented by trees. 
Each tree consists of nodes and leaves, as well as the discrete labels on the nodes (mathematical operators, e.g., $+,-,*,/$) and leaves (variables and real-valued constants). 
Figure~\ref{fig:bilevel} shows the tree structure of the symbolic expression  $\sqrt{c_1-z_2} + (z_1 c_2)$.
Our goal is to develop models that learn the relation between discrete tree structures and their fitness. For now, we are not interested in the influence of the real-valued constants. Hence, we suggest a bi-level problem definition.

\subsection{Problem Definition}
\begin{figure}[!h]
\centering
\includegraphics[width=1\textwidth]{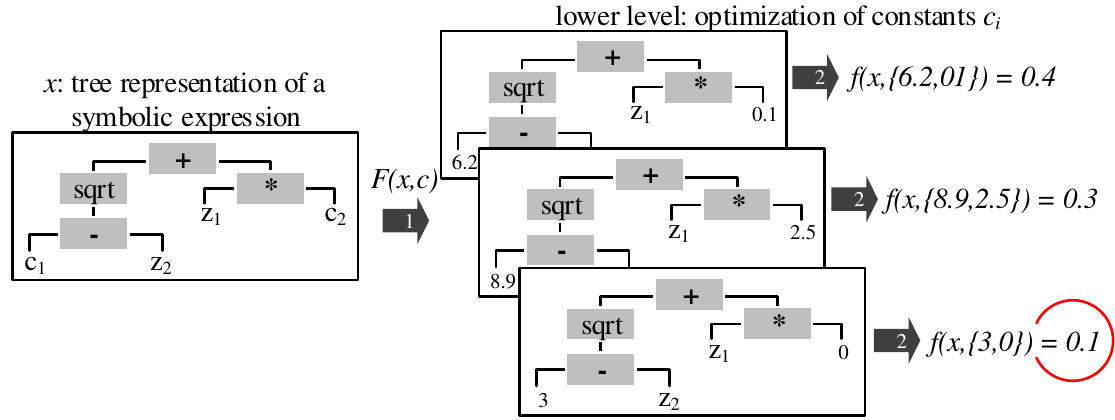}
\caption{
Example for the upper level candidate $x=\sqrt{c_1-z_2} + (z_1 c_2)$.
To estimate its fitness $F(x,c)$, a lower level optimizer (step 1)
 estimates the fitness $f(x,c)$ for different constants (step 2) and returns
 the best to $F(x,c)$ (red circle). 
}
\label{fig:bilevel}
\end{figure}

The upper level is the optimization of the discrete tree structure. For each fitness evaluation of the upper level, the lower level optimization problem has to be solved, which
comprehends the optimization of the constants.
Therefore, the upper level problem is defined by
\begin{align*}
\min_x  F(x,c) \qquad 
\text{subject to~~~} & c \in \argmin_c f(x,c),
\end{align*}
where x is the tree structure representation, 
$c \in \mathbb{R}^d$ is the set of $d_c$ constant values,
and $f(x,c)$ is the lower level objective function.
Note, that the number of constants $d_c$ depends on $x$. In extreme cases, the tree $x$ may not
contain any constants ($d_c = 0$), which eliminates the lower level problem.
The fitness will be determined as 
\begin{equation}\label{eq:fitness}
f(x,c) = 1-|\text{cor}(\hat{y}(x,c),y)|,
\end{equation}
where $\hat{y}(x,c)$ denotes the output of the symbolic expression for the data set, $y$ is the corresponding vector of true observations, and
$\cor(\cdot, \cdot)$ is the Pearson correlation coefficient.
If $\hat{y}(x,c)$ becomes infeasible (e.g., due to a negative square root or division by zero), we assign a penalty value. To that end, we use the upper bound of our fitness function, $f_{\text{penalty}}(x,c) = 1$.
An example of an upper level candidate's evaluation is visualized in Fig.~\ref{fig:bilevel}.
If not stated otherwise, fitness evaluations refer to evaluations of the upper level function $F$.

\subsection{Surrogate Model-Based Optimization}
The SMBO approach we employ for the upper level optimization is loosely based on the EGO algorithm~\cite{Jones1998}. 
Initially, the search space is randomly sampled. 
The resulting data is used to learn a suitable regression model.
This surrogate model is subject to a search via an optimization algorithm (e.g., an EA), which optimizes an infill criterion based on the model. An iteration ends with evaluating the actual (upper level) fitness of the new individual. 
Then, the surrogate model is updated with the new data and the procedure iterates.

As in standard EGO, we utilize a Kriging regression model, which assumes that the observed data is 
derived from a Gaussian process~\cite{Forrester2008a}.
One reason for the popularity of
Kriging in SMBO is that it allows to estimate
its own uncertainty. The uncertainty estimate can be used
to calculate the expected improvement (EI) infill criterion, which allows to balance exploitation and exploration in an optimization process~\cite{Mockus1978,Jones1998}.

Importantly, Kriging is based on correlation measures or kernels, which
describe the similarity of samples. Exponential kernels, e.g., $\text{k}(x,x')=\exp(-\theta ||x-x'||_2)$,
 with the parameter $\theta$ determined by Maximum Likelihood Estimation (MLE), are often used.
It is straightforward to extend kernel-based models 
to combinatorial search spaces~\cite{Moraglio2011,Zaefferer2014b}.
The core idea is to replace the distance measure, e.g., in the exponential kernel $\text{k}(x,x')=\exp(-\theta \text{d}(x,x'))$.
The distance measure $\text{d}(x,x')$ can be some adequate measure of distance
between candidate solutions, such as an edit distance.
Our study follows this idea. We will compare different distance measures and test how much they can contribute to Kriging models in an SMBO algorithm.
 
\section{Kernels for Bi-level Symbolic Regression}
We investigate four distance measures between trees or symbolic expressions, that will embedded into an exponential kernel.
\subsection{Phenotypic Distance}
The Phenotypic Distance (PhD) estimates the dissimilarity of two individuals (trees) based on their program output / phenotype, instead of using their code / genotype.
This idea has been suggested by Hildebrandt and Branke for evolving dispatching rules via GP~\cite{Hildebrandt2014}. 
They defined a phenotypic dissimilarity by comparing the outcome of a decision rule based on a small set of test situations.
Our SR tasks require a different definition of the phenotypic distance.
We propose to measure the correlation between the outcomes of 
two symbolic expressions, with all numeric constants set to one.
Hence, we save the effort of the optimization of the constants and 
compare the outputs of the expressions  $\hat{y}(x,\bm{1})$ with
\begin{equation*}
\fdistPhD(x,x') = 1-|\text{cor}(\hat{y}(x,\bm{1}),\hat{y}(x',\bm{1}))|.
\end{equation*}
If either of the two expressions is infeasible (e.g., due to division by zero), the distance will be set to one. 
Setting all constants to one is of course arbitrary. 
A random sample would also be possible but potentially problematic. A difference in phenotype could be perceived due to a different assignment of the constants on the leaves, rather than an actually different behavior of the symbolic expressions.


\subsection{Tree Edit Distance}
As an alternative to the PhD, we will also employ genotypic distances,
i.e., distances between trees. 
One possible definition of distance between trees is the minimal number of edit operations required to transform one tree into another. 
This approach is denoted as the Tree Edit Distance (TED).
We use the TED implementation that was introduced by
Pawlik and Augsten~\cite{Pawlik2016}. It is
available in the APTED library version 0.1.1~\cite{Pawlik2017}.
The APTED implementation counts the following edit operations:
node deletion, node insertion, and node relabeling.


\subsection{Structural Hamming Distance}
The Structural Hamming Distance (SHD)~\cite{Moraglio2005}
 has been used to express
genotypic dissimilarity for model-based GP in several studies~\cite{Moraglio2011b,Kattan2014,Hildebrandt2014}. 
Roughly speaking, it compares two trees by recursively checking each node
that the two trees have in common. To compare nodes, it uses the Hamming Distance (HD), which is one if two labels are different and zero otherwise. The original SHD (SHD1) is defined as 
\begin{equation*}
\fdistSHDa(x,x') =
\left\{
\begin{array}{ll}
              1, & \text{ if } \ary(x_0) \neq \ary(x'_0) \\
  \HD(x_0,x'_0), & \text{ if } \ary(x_0) = \ary(x'_0) = 0 \\
  \Delta(x,x'), & \text{ if } \ary(x_0) = \ary(x'_0) = m, \\
\end{array}
\right.
\end{equation*}
with  
\begin{equation}\label{eq:HD1}
\Delta(x,x') = \frac{1}{m+1}\left(\HD(x_0,x'_0)+\sum^m_{i=1}\fdistSHDa(x_i,x_i')\right).
\end{equation}
Here, $x$ and $x'$ are trees, $x_0$ indicates a root node of $x$, $x_i$ with $i\geq1$ is the $i$-th subtree of x, and $\ary(x_0)$ implies the number of subtrees linked to the corresponding node.
We use a slight variation, which we refer to as SHD2.
For the sake of simplicity, we define it for trees with  a maximum arity of two.
SHD1 and SHD2 are identical, except for the case
$\ary(x_0) = \ary(x'_0) = m > 1$. Then,  eq.~\eqref{eq:HD1} becomes 
\begin{align*}
\Delta(x,x')  = & \frac{1}{m+1} \Bigl(\HD(x_0,x'_0) + \\
     & \min\left\{
\fdistSHDb(x_1,x_1') + \fdistSHDb(x_2,x_2'),
\fdistSHDb(x_1,x_2') + \fdistSHDb(x_2,x_1')
\right\}
\Bigr)
.
\end{align*}
That means, when two subtrees $x_1, x_2$ are compared with their counterparts $x_1', x_2'$, we use the pairing or alignment between $x$ and $x'$ which yields the smaller distance. Potentially, this is more accurate, 
since it does not depend on the (arbitrary) initial alignment
of the two trees. But SHD2 requires additional computational effort, even more so for larger arities.

The reason for using this modified variant lies in the nature of our SMBO algorithm. 
SAEAs yield datasets where some individuals will have common ancestors (or are ancestors of each other),
and hence, are inherently more likely to be aligned
with each other.
Contrarily, SMBO generates new trees via a randomly initialized search that 
avoids direct ancestor relationships among individuals.
This implies that two trees are more likely
to have different alignments.
Then, SHD2 is a potentially more accurate (but costly) measure.

\subsection{Comparison and Linear Combination of Distances}

For the comparison of the four different distance measures, we first calculated the distance matrices for 100 randomly 
generated trees (symbolic expressions). We used the same random tree-generation method as 
in Sec.~\ref{sec:case}.
We computed the Pearson correlation between the different distance matrices.
For this sample, the SHD variants yielded a strong correlation of $0.99$, which
indicates that they reflect very similar information.
For the remaining samples, the correlation was $0.51$ (PhD, SHD2), $0.29$ (PhD, TED),
and $0.37$ (TED, SHD2). That is, the largest diversity was observed between PhD and TED.
\begin{figure}[tb]
\centering
\includegraphics[width=\textwidth]{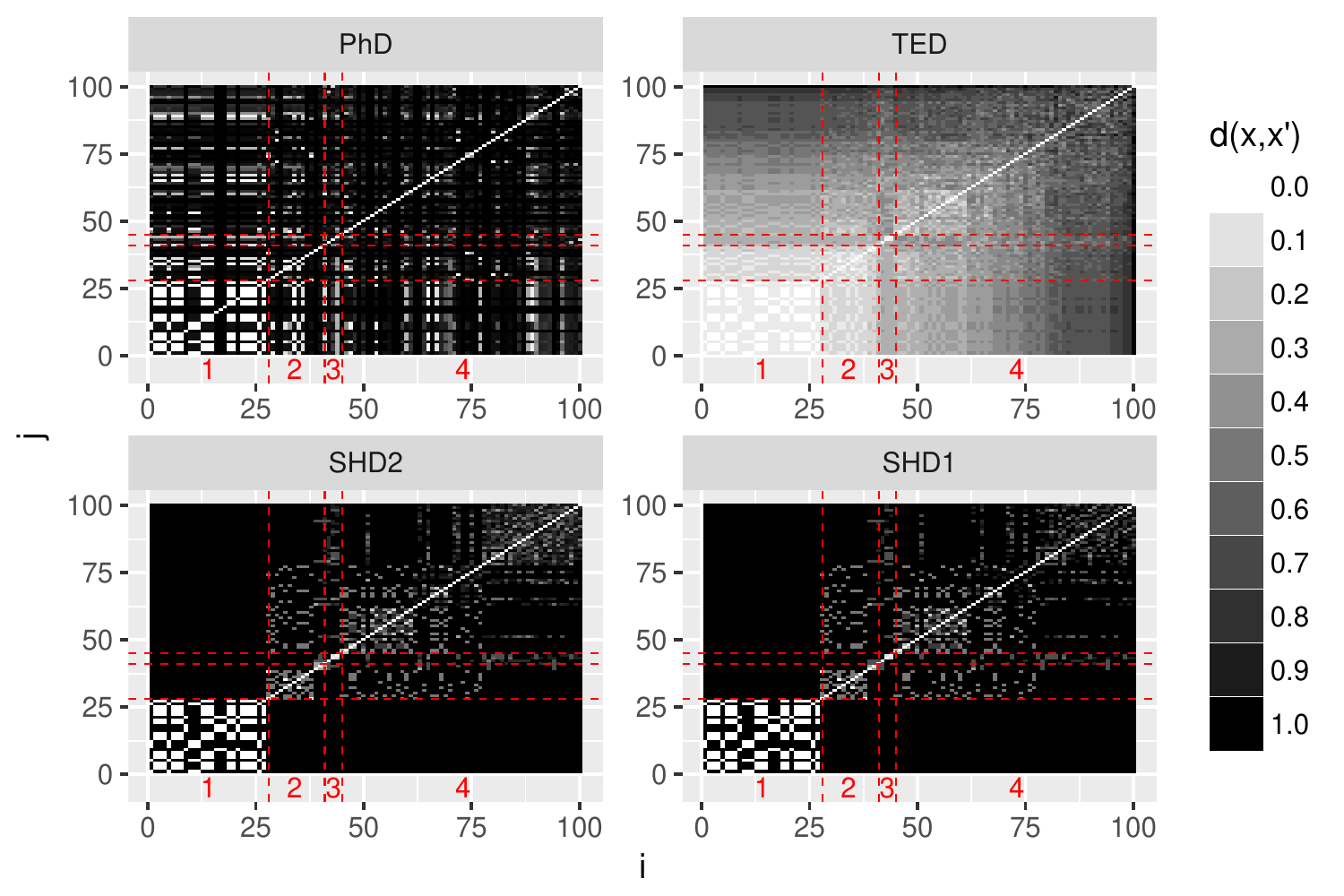}
\caption{Image plot of the four different tree distance measures. Each image cell is an element of a distance matrix.
The trees are sorted by their complexity (tree depth and number of nodes).
Trees in the lower left corner are less complex than those in the upper right. 
The tree depth is annotated in red at the bottom of each plot.
}
\label{fig:image}
\end{figure}
Figure~\ref{fig:image}
visualizes the corresponding distance matrices. 
It shows that the SHD does have problems with differentiating
between trees of different complexity. Several large blocks of the SHD matrices have a value of one, indicating
that the respective trees are at maximum distance. This lack of 
perceiving a more fine-grained difference
is problematic. It implies that any model based on SHD is potentially inaccurate
for trees of a complexity that has not been observed so far.
TED and PhD tend to see larger distances for more complex trees. 
This is obvious for TED, as complex trees require more operations to be transformed into each other.
For PhD it is clear that complex trees can produce more diverse phenotypic behavior.

With regards to the computational effort, we note that TED is by far the
most expensive measure. It is followed by the PhD and the cheapest measure is SHD1.
While the specifics strongly depend on the implementation, we note that the TED required at least an order of magnitude more computation time than the others.
This is not surprising, as determining the minimal number of edit operations requires to solve an optimization problem.

The PhD measure seems most promising in terms of generalizability. Most GP problems involve some  phenotypic behavior that may be measured/compared. SHD and TED are limited to problems with tree structures and discrete labels.

The diversity of the different distances suggests that it is promising to combine them.
We propose a linear combination
of the PhD, TED, and SHD2.
We decided to focus on one of the SHD variants due to their similarity
and chose the SHD2 variant due to its potentially
increased accuracy. Also, its increased computational cost disappears compared
to the larger costs of the TED.
The linear combination in the kernel is
\begin{align}\label{eq:lincom}
\fkern(x,x')=\exp\left\{
-\beta_1 \fdistSHDb(x,x')  
-\beta_2 \fdistPhD(x,x')  
-\beta_3 \fdistTED(x,x') 
\right\}. 
\end{align}
Each distance receives a weight  
$\beta_i \in \mathbb{R}^+$
that is determined by MLE.
The linear combination allows for a potentially more accurate Kriging model.
As we do not know a-priori which distance measure is appropriate for
a certain problem (or whether they complement each other), the combination
shifts this decision problem to the model.
Furthermore, the weights provide insights into when and how much each distance contributes to the model.

\section{Case Study}\label{sec:case}
We performed a case study, testing the SMBO algorithm with six SR tasks.

\textit{Symbolic Regression Test Problems:}
We chose the Newton, sine-cosine, Kotan\-chek2D, and Salustowicz1D problems as used in~\cite{Flasch2015b} and the sqr and sqr+log problem as used in~\cite{Kattan2014}. 
All problem configurations remained unchanged, i.e., operator set, data set size, and bounds for variables.
We did not evaluate the derived symbolic expressions
on an additional test set since our goal was to determine the ability of the SMBO algorithm
to learn the connection between candidate solutions and fitness.

\textit{Lower level optimization of the constants:} 
To optimize the lower level objective function, we decided to use
the locally biased version of the
the Dividing RECTangles (DIRECT) algorithm~\cite{Gablonsky2001} 
for a global search. DIRECT uses
 $1000 \times d_c$ evaluations of the objective function.
The result of the DIRECT run is further refined with a Nelder-Mead local search~\cite{Nelder1965} (also $1000 \times d_c$ evaluations).

\textit{Upper level optimization of the structure:}
All algorithms received a budget of 100 upper-level objective function evaluations to emulate an expensive optimization problem.
We used Random Search (RS) and a model-free EA as baselines.
All operators were taken from the \texttt{rgp} package~\cite{Flasch2014a}.
For creating new individuals, both baselines used
\texttt{randfunc\-Ramped\-Half\-And\-Half}, parameterized
with a maximum tree depth of $4$ and a probability to generate constants of $0.2$.
Furthermore, the EA employed \texttt{crossover\-expr\-Fast} for recombination, which randomly exchanges subtrees.
For mutation, \texttt{mutate\-Subtree\-Fast} was used.
The parameters of the mutation operator are as follows: 
$0.1$ (probability to insert a subtree),
$0.1$ (probability to delete a subtree),
$0.1$ (probability of creating a subtree instead of a leaf),
$0.2$ (constant generation probability), and 
$4$ (maximum tree depth).
Since constant values were not considered at the upper level, the respective bounds in the operator are both set to one.
We employed a standard EA (based on \texttt{optimEA} in the \texttt{CEGO} package~\cite{CEGOv2.2.0}) 
that used the above described operators.
The EA used truncation selection, and a fixed number of children in each generation.
The population size and number of children were tuned (see Sec.~\ref{sec:tuning}).

The upper level problem was also solved by the SMBO algorithm. 
We used the Kriging model from the \texttt{CEGO} package,
with the kernel given in eq.~\eqref{eq:lincom}.
The model was trained within $1,000$ likelihood evaluations (via DIRECT).
The EA searched on the surrogate model with $10,000$
evaluations of the EI criterion in each iteration.
The SMBO search was initialized with $20$ random trees.

For the analysis, we recorded the best individual for each run. In addition, we
recorded the weights used for linear combination of the distances in each iteration, to evaluate
the contribution of each distance function over time.
Each algorithm run was repeated $20$ times.

\subsection{Algorithm Tuning}\label{sec:tuning}
We decided to tune some potentially sensitive parameters
to allow for a more fair comparison between the model-based
and model-free algorithm.
The model-free GP algorithm's population size $\mu$ and number of 
children $\lambda$ produced in each iteration were tuned. 
All combinations of $\mu=\{5, 10, 15, 20\}$ and $\lambda = \{1,2,3,4,5\}$
were tested. The optimization
performance was expected to be sensitive to these parameters,
due to the extremely small fitness evaluation budget.

For the SMBO algorithm, we did not tune $\mu$ and $\lambda$.
Due to the overall larger complexity we decided to set the parameters
based on experience only, without a detailed tuning.
In fact, due to the larger number of evaluations (of the surrogate model)
the algorithm should be less sensitive to $\mu$ and related
parameters. Since $10,000$ evaluations of the surrogate model were allowed,
a (relative to the model-free EA) large $\mu=200$ was given
to the EA and correspondingly larger $\lambda = 10$.

We also performed preliminary experiments with the mean square error (MSE) instead
of the correlation-based fitness measurement in equation~\eqref{eq:fitness}. 
The MSE-based experiments 
yielded rather poor results with SMBO. 
This may be explained by the penalty for infeasible candidates. 
The penalty value is very
difficult to set for the MSE case. A poor choice 
may severely impair the ability to train a good Kriging model because
of strong jumps or plateaus in the fitness landscape.
While our preliminary experiments were not very detailed, they can be counted as additional tuning effort, since they influenced the choice of the correlation measure used in the phenotypic distance.

\subsection{Analysis and Discussion}
\begin{figure}[tb]
\centering
\includegraphics[width=\textwidth]{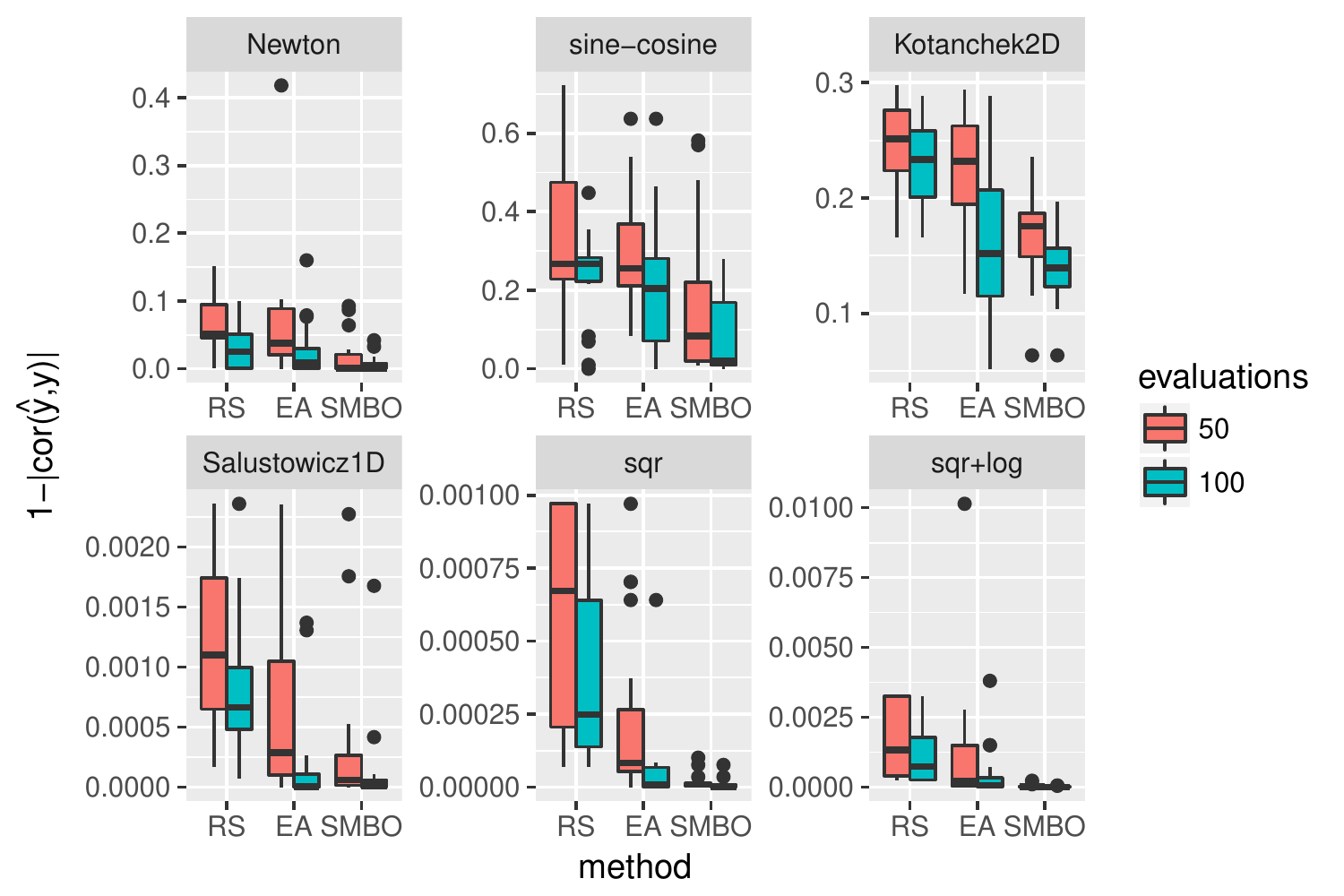}
\caption{Boxplot of best found values after 50 and 100 evaluations respectively.}
\label{fig:box}
\end{figure}

Boxplots of the best observed fitness after 50 and 100 evaluations of the objective function $F$
are shown in Fig.~\ref{fig:box}.
We report results of the tuned, model-free EA that achieved
the best mean rank on all problems ($\mu = 15$, $\lambda=1$). 
The minimal $\lambda$ makes
sense, as it allows to perform a large number of iterations despite the small budget.
For each problem and number of evaluations, we tested for statistical significance of the observed differences
via the non-parametric Kruskal-Wallis rank sum test and Conover posthoc test, with a significance level of $0.05$.
The SMBO was significantly better 
than its two competitors in most cases, except for Salustowicz1D and Kotanchek2D after $100$ evaluations, where no evidence for significant differences to the model-free EA is found.
The EA was significantly better than the plain RS, except for Newton and sine-cosine ($50$ and $100$ evaluations) as well as Kotanchek2D (50 evaluations).

\begin{figure}[tb]
\centering
\includegraphics[width=0.8\textwidth]{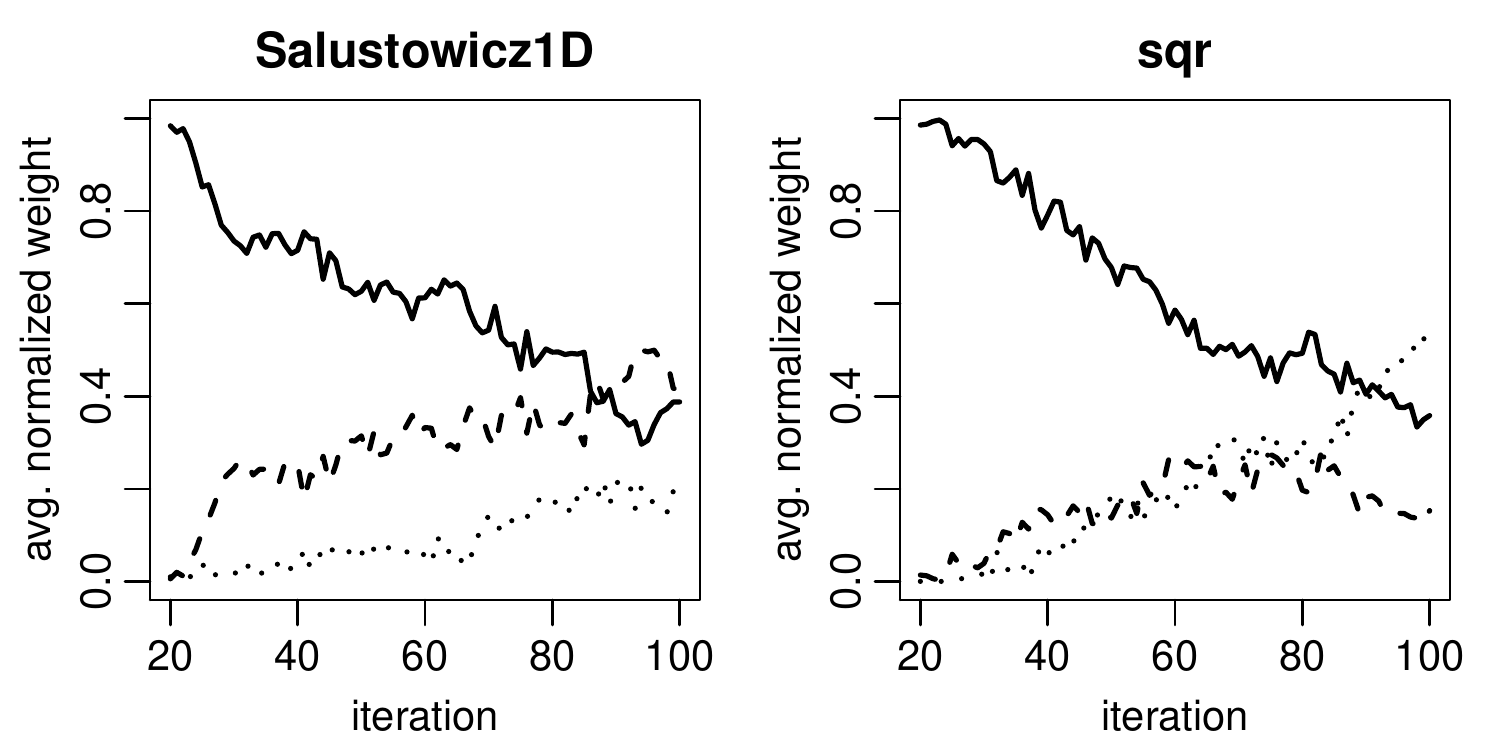}
\caption{Average normalized weights for the different kernels/distances. Solid line: PhD, dashed line: TED, dotted line: SHD2.}
\label{fig:w}
\end{figure}

To determine which distance measures contributed to these results,
the weights of the linear combination are shown in Fig.~\ref{fig:w}.
The weights are normalized so that they sum up to one.
We show results for two problems, since they are similar in the other four cases.
Usually, the PhD received the largest weights in the beginning, whereas
the importance of the TED increased throughout the run, sometimes overtaking
the PhD.   SHD usually does not contribute as much, except for the sqr problem instance. 
Here, SHD overtakes both other distances at the end of the run.
The generally larger importance of the PhD compared to SHD is in agreement with previous results by Hildebrandt and Branke~\cite{Hildebrandt2014}, 
where a similar distance achieved better results than SHD.

We confirmed these results by additional optimization experiments for each single distance (i.e., without a linear combination).
Runs with PhD tended to suggest good candidate solutions early, whereas TED and SHD performed better later on. 
The linear combination performed at least as well as the best of the single-distance models.

\section{Conclusion and Outlook}
We investigated whether
three distance measures can be employed in an SMBO algorithm based on
a Kriging model. 
We tested the algorithm with SR tasks. 
With respect to the research questions stated in Sec.~\ref{sec:intro}, our results can be summarized as follows:
\begin{compactenum}
\item The distance measures PhD, SHD and TED are quite diverse. 
The SHD differentiates poorly
between trees with different complexities.
Especially the TED seems to be much more fine grained, 
but it requires the most computational effort. On the other hand, the PhD 
is comparatively cheap to evaluate and independent of the genotype.
\item Interestingly, the PhD seemed to contribute most, followed by
the TED. This was especially true for small data sets at the beginning
of an optimization run. Later on, TED and to a lesser extent SHD gained importance.
\item A Kriging model based on
a linear combination of the three distances seems to be beneficial 
for SMBO. The SMBO algorithm outperformed a model-free algorithm and random search.
All algorithms used no more than 100 fitness evaluations.
\end{compactenum} 
In future work, we would like to determine how well these results apply to other problem classes. 
Furthermore, alternatives to the linear combination of
distances should be investigated. 

\bibliographystyle{splncs}
\bibliography{Zaef18b}  

\begin{thebibliography}{10}

\bibitem{koza1994genetic}
Koza, J.R.:
\newblock Genetic programming as a means for programming computers by natural
  selection.
\newblock Statistics and computing \textbf{4}(2) (1994)  87--112

\bibitem{Flasch2015b}
Flasch, O.:
\newblock A modular genetic programming system.
\newblock PhD thesis, TU Dortmund (2015)

\bibitem{Nguyen2017}
Nguyen, S., Mei, Y., Zhang, M.:
\newblock Genetic programming for production scheduling: a survey with a
  unified framework.
\newblock Complex {\&} Intelligent Systems \textbf{3}(1) (feb 2017)  41--66

\bibitem{Bart16n}
Bartz-Beielstein, T., Zaefferer, M.:
\newblock Model-based methods for continuous and discrete global optimization.
\newblock Applied Soft Computing \textbf{55} (2017)  154 -- 167

\bibitem{Parisotto2016}
{Parisotto}, E., {Mohamed}, A., {Singh}, R., {Li}, L., {Zhou}, D., {Kohli}, P.:
\newblock Neuro-symbolic program synthesis.
\newblock ArXiv e-prints \textbf{1611.01855} (November 2016)

\bibitem{Moraglio2011}
Moraglio, A., Kattan, A.:
\newblock Geometric generalisation of surrogate model based optimisation to
  combinatorial spaces.
\newblock In: Proceedings of the 11th European Conference on Evolutionary
  Computation in Combinatorial Optimization. EvoCOP'11, Berlin, Heidelberg,
  Germany, Springer (2011)  142--154

\bibitem{Zaefferer2014b}
Zaefferer, M., Stork, J., Friese, M., Fischbach, A., Naujoks, B.,
  Bartz-Beielstein, T.:
\newblock Efficient global optimization for combinatorial problems.
\newblock In: Proceedings of the 2014 Genetic and Evolutionary Computation
  Conference. GECCO '14, New York, NY, USA, ACM (2014)  871--878

\bibitem{Jones1998}
Jones, D.R., Schonlau, M., Welch, W.J.:
\newblock Efficient global optimization of expensive black-box functions.
\newblock Journal of Global Optimization \textbf{13}(4) (1998)  455--492

\bibitem{Jin2011}
Jin, Y.:
\newblock Surrogate-assisted evolutionary computation: Recent advances and
  future challenges.
\newblock Swarm and Evolutionary Computation \textbf{1}(2) (2011)  61--70

\bibitem{Kattan2014}
Kattan, A., Ong, Y.S.:
\newblock Surrogate genetic programming: A semantic aware evolutionary search.
\newblock Information Sciences \textbf{296} (2015)  345--359

\bibitem{Hildebrandt2014}
Hildebrandt, T., Branke, J.:
\newblock On using surrogates with genetic programming.
\newblock Evolutionary Computation \textbf{23}(3) (Jun 2015)  343--367

\bibitem{Nguyen2014}
Nguyen, S., Zhang, M., Johnston, M., Tan, K.C.:
\newblock Selection schemes in surrogate-assisted genetic programming for job
  shop scheduling.
\newblock In: Simulated Evolution and Learning, 10th International Conference,
  SEAL, Springer Science + Business Media (2014)  656--667

\bibitem{Nguyen2016}
Nguyen, S., Zhang, M., Tan, K.C.:
\newblock Surrogate-assisted genetic programming with simplified models for
  automated design of dispatching rules.
\newblock {IEEE} Transactions on Cybernetics (2016)  1--15

\bibitem{Moraglio2011b}
Moraglio, A., Kattan, A.:
\newblock Geometric surrogate model based optimisation for genetic programming:
  Initial experiments.
\newblock Technical report, University of Birmingham (2011)

\bibitem{Forrester2008a}
Forrester, A., Sobester, A., Keane, A.:
\newblock Engineering Design via Surrogate Modelling.
\newblock Wiley (2008)

\bibitem{Mockus1978}
Mockus, J., Tiesis, V., Zilinskas, A.:
\newblock The application of Bayesian methods for seeking the extremum.
\newblock In: Towards Global Optimization 2. North-Holland (1978)  117--129

\bibitem{Pawlik2016}
Pawlik, M., Augsten, N.:
\newblock Tree edit distance: Robust and memory-efficient.
\newblock Information Systems \textbf{56} (mar 2016)  157--173

\bibitem{Pawlik2017}
Pawlik, M., Augsten, N.:
\newblock APTED release 0.1.1.
\newblock GitHub. (2016) \url{https://github.com/DatabaseGroup/apted}, last
  accessed: 2017-06-01.

\bibitem{Moraglio2005}
Moraglio, A., Poli, R.:
\newblock Geometric landscape of homologous crossover for syntactic trees.
\newblock In: 2005 {IEEE} Congress on Evolutionary Computation, Edinburgh, UK,
  {IEEE} (2005)

\bibitem{Gablonsky2001}
Gablonsky, J., Kelley, C.:
\newblock A locally-biased form of the direct algorithm.
\newblock Journal of Global Optimization \textbf{21}(1) (September 2001)
  27--37

\bibitem{Nelder1965}
Nelder, J.A., Mead, R.:
\newblock A simplex method for function minimization.
\newblock The Computer Journal \textbf{7}(4) (Jan 1965)  308--313

\bibitem{Flasch2014a}
Flasch, O., Mersmann, O., Bartz-Beielstein, T., Stork, J., Zaefferer, M.:
\newblock rgp: R genetic programming framework. (2014) R package version 0.4-1.

\bibitem{CEGOv2.2.0}
Zaefferer, M.:
\newblock Combinatorial efficient global optimization in {R} - {CEGO} v2.2.0.
\newblock online: https://cran.r-project.org/package=CEGO (2017) accessed:
  2018-01-10.

\end{thebibliography}

\end{document}